\documentclass[letterpaper]{article} 
\usepackage[preprint]{aaai2027} 
\usepackage[hyphens]{url} 
\usepackage{graphicx} 
\usepackage{natbib} 
\usepackage{caption} 
\usepackage{amsmath,amssymb}
\usepackage{multirow}

\title{FutureBridge: Token Selection Beyond Local Preference in Collaborative Decoding}
\author{
Quanquan Li\equalcontrib\textsuperscript{\rm 1},
Hongbo Zhang\equalcontrib\textsuperscript{\rm 2},
Yihe Chi\textsuperscript{\rm 1},
Jingyu Li\textsuperscript{\rm 3},
Xidong Xi\textsuperscript{\rm 1},
Liuyang Song\textsuperscript{\rm 2},\\
Hongzhen Zhang\textsuperscript{\rm 1},
Yuxiang Huang\textsuperscript{\rm 1},
Jing Ke\textsuperscript{\rm 5},
Siyuan Ma\textsuperscript{\rm 4},
Junyi Lin\textsuperscript{\rm 6},
Guitao Cao\textsuperscript{\rm 1}\corresponding
}
\affiliations{
\textsuperscript{\rm 1}East China Normal University\\
\textsuperscript{\rm 2}Peking University\\
\textsuperscript{\rm 3}University of Science and Technology of China\\
\textsuperscript{\rm 4}Nanyang Technological University\\
\textsuperscript{\rm 5}Shanghai Jiao Tong University\\
\textsuperscript{\rm 6}Guangdong University of Technology
}

\begin{document}
\maketitle

\begin{abstract}
Token-level collaboration allows a large language model (LLM) to assist a small language model (SLM) when their predictions diverge. Existing methods either use LLM-generated intervention tokens or rank candidates with the LLM's next-token probabilities. Both rely on the LLM's local preference, even though an LLM-selected token may be difficult for the SLM to build on. We present FutureBridge, which ranks joint LLM--SLM token candidates according to how well they support the SLM's subsequent reasoning. During training, an answer-verified LLM trajectory supplies a fixed shared future, and a frozen SLM evaluates every candidate under this common context. The resulting counterfactual scores supervise a lightweight token reranker that observes only the current state and candidate token. At inference, FutureBridge uses the LLM only to expand the candidate pool, selects one token, and returns generation to the SLM without generating or appending a future suffix. Across five mathematical reasoning benchmarks, FutureBridge improves the Qwen3-1.7B SLM's Math Avg. by 35.1\% relative to greedy SLM decoding. These results indicate that token selection benefits from modeling whether the receiving SLM can use each candidate to continue reasoning, rather than relying on the LLM's local preference alone.
\end{abstract}

\section{Introduction}

Token-level small--large model collaboration provides a fine-grained trade-off between reasoning capability and computational cost by invoking the LLM on demand during generation~\cite{ref2,ref3,ref10}. Rather than assigning an entire request to the LLM, the SLM generates independently for most steps and requests local assistance only at selected reasoning states. Mathematical reasoning, however, is highly sensitive to local decisions: a token that is unsuitable for the SLM can redirect the subsequent trajectory away from a correct solution. For example, the LLM may prefer a compressed reasoning transition, whereas a less capable SLM may require an explicit intermediate step to continue the derivation reliably. Effective token-level collaboration therefore depends not only on when the LLM is invoked, but also on whether the provided local content can be effectively used by the SLM.

Existing small--large model collaboration methods primarily determine whether, when, and for how long to invoke the LLM~\cite{ref1,ref2,ref3,ref4,ref5,ref6,ref7,ref10,ref11,ref12}. Once collaboration is activated, some methods directly adopt LLM-generated intervention tokens, while others rank candidates using the LLM's local probabilities. S2T asks the LLM to rerank the SLM's top-$K$ candidates and distills the resulting ranking into a local selector~\cite{ref13}. This design relies on two assumptions: a useful action is already present in the SLM candidate set, and the action locally preferred by the LLM is also suitable for the SLM to continue reasoning. The former produces a candidate coverage gap when a useful LLM candidate falls outside the SLM candidate set. The latter produces a teaching suitability gap when the LLM favors a compressed reasoning transition but the SLM requires an explicit intermediate step. Local LLM preference therefore does not provide sufficient evidence that the SLM can effectively use the selected token.

These two gaps fundamentally arise from the absence of candidate-level supervision. An ideal selection signal would compare the complete reasoning outcomes produced by the same SLM from different candidates, but executing one rollout per candidate incurs substantial computational cost. A standard teacher trajectory observes only the teacher action and its corresponding future, while generating a separate future for every candidate changes both the candidate action and the evaluation context, preventing a controlled comparison. The central challenge is therefore to construct SLM-conditioned supervision that is directly comparable across candidates under a fixed downstream context.

To construct such supervision, FutureBridge uses a reasoning suffix extracted from an answer-verified LLM trajectory as a common reference for all candidates. This choice is motivated by the intuition that local generation probabilities alone are insufficient to identify which candidate better supports the SLM's subsequent reasoning; evaluating different candidates against the same answer-consistent continuation provides downstream information for the current token decision. During training, FutureBridge holds the current reasoning state and the suffix fixed and changes only the token at the current position across branches. Because the SLM's top-$K$ set may omit a critical token required for subsequent reasoning, we augment it with LLM candidates. For each token in the joint candidate set, the frozen SLM computes the average log-likelihood of the shared suffix conditioned on that candidate and uses it as a proxy for compatibility between the candidate and subsequent reasoning. In this way, one answer-consistent continuation provides dense and directly comparable supervision for the entire candidate set.

\begin{figure*}[t]
\centering
\includegraphics[width=\textwidth]{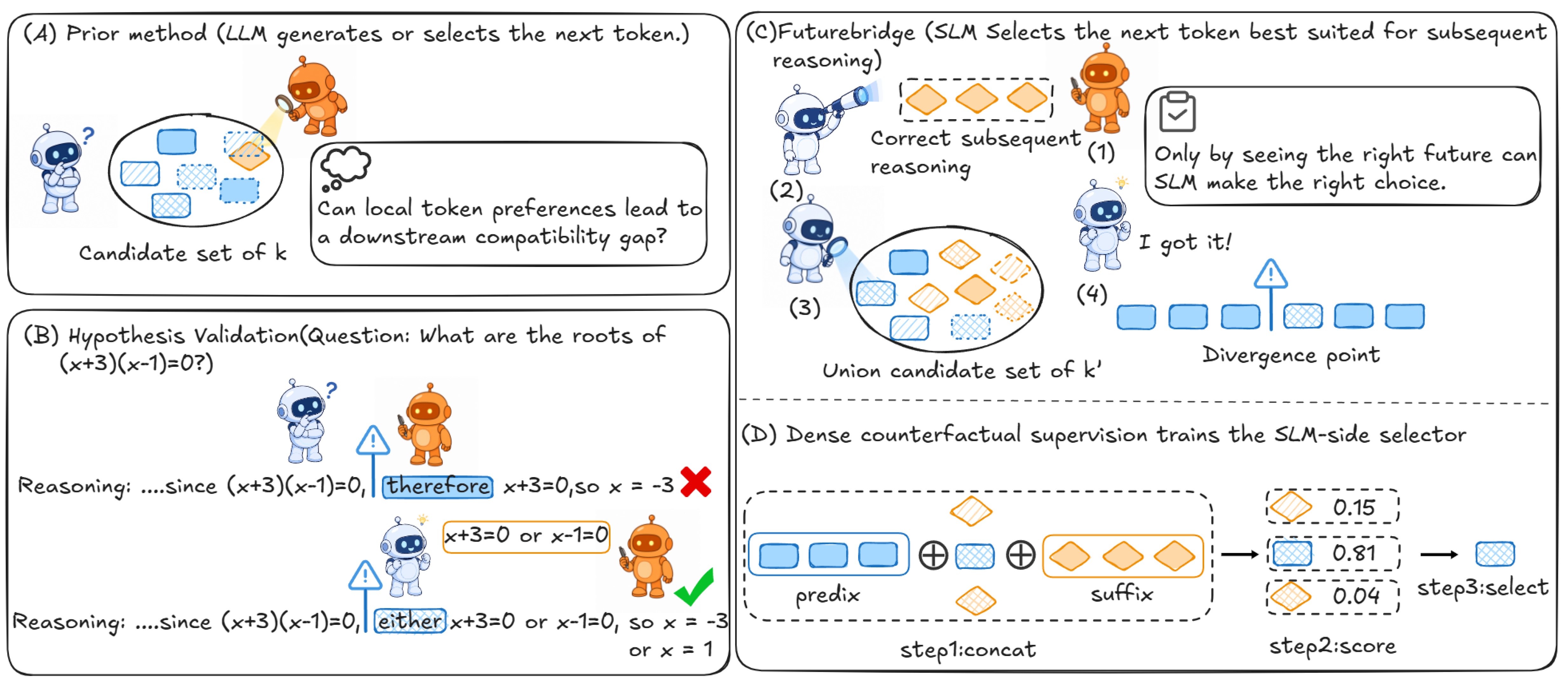}
\caption{Overview of FutureBridge. (A) Prior token-level collaboration selects interventions using the LLM's local preferences. (B) A motivating example illustrates that locally plausible tokens can differ in how well they support the SLM's subsequent reasoning. (C) FutureBridge constructs a joint candidate pool for token reranking. (D) During training, every candidate is scored against the same answer-verified future, and the resulting supervision is distilled into a token reranker that does not observe the future at inference.}
\label{fig:overview}
\end{figure*}

Building on this candidate-level supervision, FutureBridge distills the future-compatibility scores into a lightweight token reranker. At inference, the SLM determines whether to request assistance under a fixed request policy. Once collaboration is triggered, the LLM contributes only next-token candidates, and the reranker conditions on the current reasoning state and each candidate to rank the joint teacher--student pool. Only the selected token is appended to the context, after which generation immediately returns to the SLM. Thus, answer-verified futures are privileged training information rather than an input or generated artifact at deployment. Under this protocol, experiments on five mathematical reasoning benchmarks show that FutureBridge consistently improves the SLM's end-to-end reasoning performance under the same request policy and candidate budget. Matched candidate-level rollout analyses further test whether the future-compatibility score identifies candidates that support the SLM's subsequent reasoning more accurately than local token probabilities, while controlled ablations isolate the contributions of candidate expansion and future-based supervision.

Our contributions are summarized as follows:

\begin{enumerate}
\item We identify a teaching suitability gap in token-level collaboration driven by the LLM's local preferences: tokens preferred by the LLM may not support the SLM's subsequent reasoning. We therefore formulate candidate selection as a local decision problem oriented toward the SLM's subsequent reasoning.
\item We propose FutureBridge, which scores a joint SLM--LLM candidate pool against an answer-verified shared future under the frozen SLM. The resulting compatibility supervision is distilled into a lightweight token reranker that neither observes nor generates the future at inference.
\item Experiments on five mathematical reasoning benchmarks demonstrate consistent end-to-end gains and show that SLM-conditioned future compatibility provides a more effective token-selection signal than local preference.
\end{enumerate}

\section{Related Work}

\subsection{Token-Level Small--Large Model Collaboration}

Small--large model collaboration has evolved from query-level routing to fine-grained intervention during generation~\cite{ref1,ref19}. CITER and R2R delegate a small set of critical tokens to the large model~\cite{ref2,ref3}, while RelayLLM trains the small model to request large-model assistance at selected positions~\cite{ref10}. Confidence-guided routing, GlimpRouter, and TrigReason further use local uncertainty or reasoning-risk signals to allocate model computation at token, reasoning-step, or event granularity~\cite{ref5,ref6,ref7}. FusionRoute additionally combines token-level expert selection with complementary logits to refine the selected model's output distribution~\cite{ref22}. These methods primarily determine when additional computation should be introduced and which model should generate. Once collaboration is activated, choosing among multiple concrete token candidates becomes a separate problem.

\subsection{Local Token Candidate Selection}

At an admitted intervention, local candidates can be verified or rescored using different signals. Speculative Decoding uses a lightweight draft model to propose tokens or short continuations that are verified by a target model, preserving the target distribution~\cite{ref20}. Reward-Guided Speculative Decoding further introduces process rewards to balance candidate quality and target-model computation~\cite{ref8}. Contrastive Decoding instead adjusts the next-token distribution using the likelihood difference between expert and amateur models~\cite{ref21}. These methods primarily serve target-model acceleration or distribution shaping.

S2T applies candidate selection to improve small-model reasoning: the large model ranks only the small model's top-$K$ candidates, and S2T-Local distills this ranking into a local selector~\cite{ref13}. However, its candidate space is restricted to small-model proposals, and its supervision still reflects teacher preference. The methods above do not explicitly evaluate candidate tokens by how well the small model can continue under a common downstream context.

\subsection{Future-Guided Collaborative Decisions}

Several methods use downstream information to improve current decisions. SpecReason and SpecCoT generate or verify future reasoning content to decide whether to accept the current reasoning unit, treating the future as an object of verification~\cite{ref9,ref14}. R2R generates separate large-model continuations from the SLM's and LLM's top-1 tokens and uses a verifier to determine whether the SLM token changes the reasoning path~\cite{ref3}. AlphaRouter learns binary small--large model routing policies through tree search and final rewards~\cite{ref4}, while Local Branch Routing expands a local candidate tree at inference time and selects the current branch from post-candidate hidden states~\cite{ref23}. In contrast, FutureBridge uses one answer-verified LLM suffix only to construct offline supervision for a joint SLM--LLM candidate pool. Holding this training suffix fixed prevents the candidate and evaluation context from changing together, while distillation removes the suffix from the deployed selector.

\section{Problem Formulation}

We formulate token-level collaboration as a token reranking problem over a joint candidate pool constructed by an SLM $M_S$ and an LLM $M_T$ that share the same tokenizer. Given a problem $x$ and an SLM-generated reasoning prefix $y_{<t}$, the current decoding state is $s_t=(x,y_{<t})$. We denote the two models' next-token distributions by $p_S(\cdot\mid s_t)$ and $p_T(\cdot\mid s_t)$, respectively, and use $\oplus$ for token-sequence concatenation. When a fixed collaboration policy requests assistance, the models provide their respective top-$k$ candidate sets, whose union is $C_t=C_t^S\cup C_t^T$. The selector chooses one token $c\in C_t$, appends only that token to the prefix, and then returns generation to the SLM. Let $G_S(s_t,c)$ denote the continuation produced by a fixed deterministic SLM decoding policy after receiving $c$, and let $R(x,\hat y)\in\{0,1\}$ indicate final-answer correctness. The ideal token decision is
\begin{equation}
c_t^*=\arg\max_{c\in C_t}
R\!\left(x,y_{<t}\oplus c
\oplus G_S(s_t,c)\right).
\label{eq:student_value}
\end{equation}
This objective differs from selecting the token with the highest local LLM probability: the best intervention is the candidate that leads to the strongest subsequent SLM outcome. Because evaluating every candidate through a complete SLM rollout is prohibitively expensive at inference time, our goal is to use privileged future information during training to learn a lightweight token reranker that approximates this student-conditioned downstream value without observing a future suffix at deployment.

\section{Method}

\subsection{Method Overview}

FutureBridge addresses token reranking over a joint candidate pool at states admitted by a fixed request policy. At decoding step $t$, the policy determines whether to invoke collaboration from the current state and the SLM distribution:
\begin{equation}
g_t=\pi_{\mathrm{req}}\!\left(s_t,p_S(\cdot\mid s_t)\right)\in\{0,1\}.
\label{eq:request_policy}
\end{equation}
Equation~\eqref{eq:request_policy} defines an external request policy shared by all candidate selectors. The policy and per-trajectory intervention budget are not optimized by FutureBridge. To construct training supervision independently of any selector, we first run the frozen SLM on the training problems and record the states admitted by the request policy. No candidate intervention or LLM suffix is inserted during this state-collection stage, so every candidate selector is trained from the same logged states.

For each state with $g_t=1$, FutureBridge proceeds in three training stages. First, the SLM and LLM construct a joint candidate pool, and a shared future is extracted from a factual LLM trajectory whose final answer is correct. Second, FutureBridge fixes the state and shared future, replaces only the current token, and uses the frozen SLM to measure candidate compatibility. Third, the resulting candidate-level targets are distilled into a lightweight token reranker that receives only the current state and candidate token. At inference, the LLM provides next-token candidates, the reranker selects one token from the joint pool, and generation returns immediately to the SLM. Figure~\ref{fig:overview} summarizes this separation between privileged training supervision and token-only deployment.

\subsection{Joint Candidate Pool and Verified Shared Future}

For each admitted state, the SLM and LLM provide candidates from their next-token distributions:

\begin{equation}
\begin{aligned}
C_t^S &= \operatorname{TopK}\!\left(p_S(\cdot\mid s_t),K_S\right),\\
C_t^T &= \operatorname{TopK}\!\left(p_T(\cdot\mid s_t),K_T\right),\\
C_t &= C_t^S\cup C_t^T,
\end{aligned}
\label{eq:joint_candidates}
\end{equation}
where $K_S$ and $K_T$ are positive integers specifying the SLM and LLM candidate budgets. Because $C_t^S$ and $C_t^T$ are sets, Equation~\eqref{eq:joint_candidates} removes duplicate tokens automatically. Restricting the pool to $C_t^S$ assumes that a useful intervention token always remains in the SLM's local shortlist. Conversely, using only $C_t^T$ discards locally plausible SLM actions and makes candidate availability entirely LLM-defined. The union separates candidate availability from candidate selection: the LLM expands the action space, while the subsequent SLM-conditioned score determines which available token is most compatible with the receiving SLM. The construction does not presume that an LLM candidate is preferable to an SLM candidate; it places both sources in a common reranking space.

Let $\mathcal H=\{16,32,64,128\}$ denote the future horizons considered for model selection, and let $H_{\max}=\max\mathcal H=128$. All horizons are constructed from prefixes of the same maximum-length shared future.

Once the joint pool is fixed, all candidates require directly comparable supervision. Evaluating Equation~\eqref{eq:student_value} would require one complete SLM rollout per candidate. These free-running trajectories have different continuations, incur cost linear in the pool size, and provide only sparse final-answer feedback. FutureBridge instead uses greedy LLM decoding to generate one complete trajectory from the same state $s_t$ and checks its final answer with the task-specific verifier $\mathcal V$:

\begin{equation}
\begin{aligned}[c]
&\bar y_{\geq t}^{T}
=a_t^{T}\oplus z_{1:\ell_t}^{T},\\
&\mathcal V\!\left(
x,y_{<t}\oplus\bar y_{\geq t}^{T}
\right)=1,\\
&f_t=z_{1:H_{\max}}^{T},\\
&f_t^{(H)}=f_{t,1:H},\qquad 1\leq H\leq H_{\max}\leq\ell_t,
\end{aligned}
\label{eq:verified_future}
\end{equation}
where $a_t^T$ is the greedy LLM token at position $t$, $z_{1:\ell_t}^T$ is its complete post-action continuation, $f_t$ is the maximum-length shared future, and $f_t^{(H)}$ is its length-$H$ prefix. Greedy decoding implies $a_t^T\in C_t^T$ for $K_T\geq1$. Equation~\eqref{eq:verified_future} retains only trajectories whose complete factual continuation is answer-correct and contains at least $H_{\max}$ post-action tokens. Verification applies to $a_t^T\oplus z_{1:\ell_t}^T$ as a complete factual continuation; it does not assert that $f_t$ remains correct after an arbitrary candidate replacement.

For each candidate $c_k\in C_t$, we remove the original LLM token $a_t^T$, insert $c_k$ at the same position, and keep the shared future $f_t$ fixed:

\begin{equation}
\tilde y_{k,\geq t}=c_k\oplus f_t.
\label{eq:counterfactual_branch}
\end{equation}

The branch with $c_k=a_t^T$ recovers the corresponding factual prefix of the LLM trajectory; every other branch is a controlled candidate replacement. All branches share the same problem $x$, SLM reasoning prefix $y_{<t}$, shared future $f_t$, and scoring model. Thus, Equation~\eqref{eq:counterfactual_branch} changes only the token inserted at position $t$ and provides a common downstream context for candidate-level compatibility.

The shared future $f_t$ is privileged information used only to construct offline training targets. It is neither generated nor provided to the reranker at inference. Consequently, deployment does not require a suffix-quality assumption and cannot inject an unverified LLM continuation into the SLM context.

\subsection{Student-Conditioned Future Compatibility}

The candidate group above differs only in the token inserted at the current position. We therefore measure candidate compatibility by the conditional likelihood that the receiving SLM assigns to the shared future. For each candidate $c_k\in C_t$, the frozen SLM evaluates the shared future under teacher forcing, and one batched forward pass produces

\begin{equation}
\begin{aligned}
\ell_{k,h}
&=\log p_S\!\left(f_{t,h}\mid s_t,c_k,f_{t,<h}\right),\\
\mathbf L
&=[\ell_{k,h}]\in\mathbb R^{|C_t|\times H_{\max}},
\end{aligned}
\label{eq:value_matrix}
\end{equation}
where $k\in\{1,\ldots,|C_t|\}$ indexes candidates and $h\in\{1,\ldots,H_{\max}\}$ indexes positions in the shared future. The quantity $\ell_{k,h}$ is the conditional log-probability that the SLM assigns to $f_{t,h}$ after inserting $c_k$ at state $s_t$ and observing the preceding shared-future prefix $f_{t,<h}$. Equation~\eqref{eq:value_matrix} collects these token-level compatibilities for all candidates and future positions.

For a given future horizon $H$, we define the student-conditioned compatibility score of candidate $c_k$ as

\begin{equation}
B_H(c_k\mid s_t,f_t^{(H)})=
\frac{1}{H}\sum_{h=1}^{H}\ell_{k,h}.
\label{eq:bridge_score}
\end{equation}

The score in Equation~\eqref{eq:bridge_score} is the average conditional log-likelihood that the receiving SLM assigns to the same length-$H$ shared future after candidate $c_k$ is inserted. Length normalization prevents the score from accumulating mechanically with suffix length and makes different horizons comparable. We write $B_H(c_k)$ when $s_t$ and $f_t^{(H)}$ are clear from context.

The reasoning prefix, candidate pool, shared future, and scoring model remain identical across candidate branches, so score differences are induced only by the current candidate. A candidate-specific LLM future would vary both the candidate and evaluation target while making LLM generation cost grow linearly with the pool size. The shared future instead evaluates every candidate under one downstream context and yields directly comparable candidate-level supervision.

Importantly, $B_H$ measures compatibility under teacher forcing rather than final-answer correctness under free-running generation. It is a tractable surrogate for the complete SLM rollout value in Equation~\eqref{eq:student_value}, not a causal outcome. We test its relationship with final reasoning outcomes using complete candidate rollouts from matched states.

To convert compatibility scores into candidate-level supervision, we standardize them within each candidate group. For any score function $r:C_t\rightarrow\mathbb R$, define
\begin{equation}
\begin{aligned}
\mu_r
&=\frac{1}{|C_t|}\sum_{c\in C_t}r(c),\\
\sigma_r
&=\sqrt{\frac{1}{|C_t|}
\sum_{c\in C_t}(r(c)-\mu_r)^2},\\
\operatorname{Norm}_{C_t}(r(c))
&=\frac{r(c)-\mu_r}{\sigma_r+\epsilon},
\end{aligned}
\label{eq:group_normalization}
\end{equation}
where $\epsilon>0$ ensures numerical stability. Equation~\eqref{eq:group_normalization} aligns the scale of scores within each candidate group. The main FutureBridge target uses only SLM-conditioned future compatibility:

\begin{equation}
q_H(c\mid s_t,f_t^{(H)},C_t)=
\frac{
\exp\!\left(\operatorname{Norm}_{C_t}(B_H(c))/\tau\right)}
{\sum_{c'\in C_t}
\exp\!\left(\operatorname{Norm}_{C_t}(B_H(c'))/\tau\right)},
\label{eq:soft_target}
\end{equation}
where $\tau>0$ is the target temperature. Equation~\eqref{eq:soft_target} encodes both the relative ordering and compatibility differences within the candidate group, providing denser supervision than a hard top-1 label.

The future horizon is a supervision and model-selection hyperparameter rather than a deployment-time generation length. We train each $H\in\mathcal H$ under the same states, candidate pools, model architecture, and training budget, then select one horizon $H^*$ and its reranker checkpoint on MATH validation. The selected checkpoint is fixed for every held-out benchmark; FutureBridge does not learn a state-dependent horizon policy or generate $H^*$ future tokens at inference.

To isolate the effect of LLM local preference, we retain the following hybrid target only as an ablation:
\begin{equation}
\begin{aligned}
U_H^{(\alpha)}(c)={}&
\alpha\operatorname{Norm}_{C_t}(B_H(c))\\
&+(1-\alpha)
\operatorname{Norm}_{C_t}\!\left(\log p_T(c\mid s_t)\right).
\end{aligned}
\label{eq:hybrid_ablation}
\end{equation}
For this ablation, Equation~\eqref{eq:hybrid_ablation} replaces the standardized compatibility term in Equation~\eqref{eq:soft_target} before temperature normalization. When $\alpha=1$, it reduces to the main FutureBridge supervision; when $\alpha=0$, it reduces to pure LLM local preference. Configurations with $\alpha<1$ are reported as supervision ablations and are not part of the core method.

\subsection{Distilling Future Compatibility for Token Reranking}

Constructing the soft target in Equation~\eqref{eq:soft_target} requires an answer-correct factual trajectory and frozen-SLM compatibility scoring, so it is restricted to offline supervision. We distill this candidate-group target into a lightweight token reranker.

For each candidate $c_k\in C_t$, a LoRA candidate scorer based on the frozen SLM receives the augmented input $(s_t,c_k)$ and produces the scalar logit

\begin{equation}
r_{\theta_H}(c_k\mid s_t)
=g_{\theta_H}(s_t,c_k)\in\mathbb R,
\label{eq:candidate_logit}
\end{equation}
where $\theta_H$ denotes the trainable parameters learned from horizon-$H$ targets and the base SLM remains frozen. Equation~\eqref{eq:candidate_logit} assigns each candidate an independent scalar score without exposing the shared future to the scorer. The predicted candidate distribution is

\begin{equation}
\hat q_{\theta_H}(c\mid s_t,C_t)
=
\frac{\exp(r_{\theta_H}(c\mid s_t))}
{\sum_{c'\in C_t}
\exp(r_{\theta_H}(c'\mid s_t))}.
\label{eq:selector_distribution}
\end{equation}

Equation~\eqref{eq:selector_distribution} makes the dependence on $C_t$ explicit while preserving independent candidate scoring before normalization. For each $H\in\mathcal H$, let $\mathcal D_{\mathrm{FB}}^{(H)}$ denote the logged states, shared futures, and candidate groups used to construct horizon-$H$ targets. All horizons use the same logged states, candidate pools, model architecture, and optimization budget. The reranker for horizon $H$ minimizes

\begin{equation}
\begin{aligned}
\mathcal L_{\mathrm{FB}}^{(H)}(\theta_H)
&=-\frac{1}{|\mathcal D_{\mathrm{FB}}^{(H)}|}
\sum_{(s_t,f_t^{(H)},C_t)\in\mathcal D_{\mathrm{FB}}^{(H)}}\\
&\quad \sum_{c\in C_t}
q_H(c\mid s_t,f_t^{(H)},C_t)
\log \hat q_{\theta_H}(c\mid s_t,C_t).
\end{aligned}
\label{eq:distillation_loss}
\end{equation}

Equation~\eqref{eq:distillation_loss} is a group-level cross-entropy objective that trains the reranker to predict the SLM-conditioned future-compatibility distribution from $(s_t,c)$ alone. MATH validation selects the supervision horizon $H^*$ and its corresponding reranker checkpoint, which are then frozen for every held-out benchmark.

At an admitted test event, the LLM provides only $C_t^T$ from its next-token distribution. After forming the joint pool $C_t$, the selected reranker evaluates every candidate from the current state and chooses
\begin{equation}
\hat c_t=\arg\max_{c\in C_t}
\hat q_{\theta_{H^*}}(c\mid s_t,C_t).
\label{eq:inference_selection}
\end{equation}
Equation~\eqref{eq:inference_selection} selects the current token without generating a shared future, generating candidate-specific futures, or executing candidate rollouts. The system appends only $\hat c_t$ and returns generation to the SLM.

The deployed selector learns the relative value of alternative local actions rather than the LLM future itself. FutureBridge therefore relies on the LLM only for candidate proposal at admitted states. The matched LLM-preference ablation uses the same logged states, intervention budget, joint candidate pool, reranker architecture, and optimization budget; only its candidate-level supervision target changes.

\begin{table*}[t]
\centering
\caption{End-to-end accuracy (\%) with Qwen3 SLM--LLM model pairs. Qwen3-1.7B is the primary SLM, and Qwen3-0.6B evaluates scale generalization.}
\label{tab:main_results}
\small
\begin{tabular*}{\textwidth}{@{\extracolsep{\fill}}cl|rrr|rrrr}
\noalign{\hrule height 1.1pt}
\textbf{Param.} & \textbf{Benchmark}
& \multicolumn{3}{c|}{\textbf{SLM-only}}
& \multicolumn{4}{c}{\textbf{LLM-involved}} \\
\cline{3-5}\cline{6-9}
& & Greedy & Maj@8 & S2T-Local
& R2R & Takeover & S2T & FutureBridge \\
\hline
\multirow{6}{*}{\textbf{1.7B}}
& GSM8K         & 82.0 & 83.5 & \textbf{86.7} & 95.3 & 95.5 & 95.8 & \textbf{96.8} \\
& MATH-500      & 64.0 & 65.0 & \textbf{67.6} & 78.4 & 79.0 & 79.7 & \textbf{81.6} \\
& OlympiadBench & 27.1 & 27.9 & \textbf{28.9} & 41.5 & 42.0 & 42.4 & \textbf{43.5} \\
& AIME 2024     & 5.6  & 6.7  & \textbf{7.8}  & 13.3 & 14.4 & 15.6 & \textbf{16.7} \\
& AIME 2025     & 7.8  & 8.9  & \textbf{10.0} & 10.0 & 11.1 & 12.2 & \textbf{13.3} \\
\cline{2-9}
& \textit{Math Avg.} & 37.30 & 38.40 & \textbf{40.20} & 47.70 & 48.40 & 49.14 & \textbf{50.38} \\
\hline
\multirow{6}{*}{\textbf{0.6B}}
& GSM8K         & 57.8 & 66.2 & \textbf{68.5} & 76.0 & 76.8 & 77.5 & \textbf{87.8} \\
& MATH-500      & 33.5 & 41.5 & \textbf{47.2} & 52.5 & 53.0 & 54.2 & \textbf{74.8} \\
& OlympiadBench & 13.2 & 18.0 & \textbf{20.5} & 27.0 & 28.0 & 28.5 & \textbf{31.2} \\
& AIME 2024     & 4.4  & 6.7  & \textbf{7.8}  & 10.0 & 11.1 & 12.2 & \textbf{13.3} \\
& AIME 2025     & 3.3  & 5.6  & \textbf{6.7}  & 7.8  & 8.9  & 10.0 & \textbf{11.1} \\
\cline{2-9}
& \textit{Math Avg.} & 22.44 & 27.60 & \textbf{30.14} & 34.66 & 35.56 & 36.48 & \textbf{43.64} \\
\noalign{\hrule height 1.1pt}
\end{tabular*}
\end{table*}

\section{Experiments}

\subsection{Experimental Setup}

\paragraph{Benchmarks and Metrics.}
We evaluate on GSM8K, MATH-500, OlympiadBench, AIME 2024, and AIME 2025. Dataset definitions and evaluation conventions follow their original sources~\cite{ref37,ref38,ref39,ref40,ref41}. The primary metric is single-trajectory pass@1 accuracy, and Math Avg.\ is the unweighted mean over the five benchmarks. The horizon ablation varies only the amount of future context used to construct offline supervision; it does not change the deployment-time generation length. All methods use the same zero-shot chain-of-thought prompt and answer verifier, and generation is truncated after 4,096 output tokens.

\paragraph{Models and Baselines.}
We use Qwen3-1.7B as the primary SLM, Qwen3-0.6B for scale generalization, and Qwen3-32B as the LLM~\cite{ref31}; all base parameters remain frozen. The main comparison retains the methods most directly related to local token selection: greedy decoding, Maj@8~\cite{ref33}, S2T-Local~\cite{ref13}, R2R~\cite{ref3}, Takeover, and S2T~\cite{ref13}. Compatible baselines use the same Qwen3 pairs. Under the S2T request schedule, Takeover delegates the remaining trajectory to the LLM at the first admitted state. Additional baselines are reported in the supplement.

\paragraph{Collaboration and Implementation Protocol.}
FutureBridge constructs supervision from MATH training and uses MATH validation to select the request threshold, reranker checkpoint, and $H\in\{16,32,64,128\}$; all choices are then fixed. Token-level collaborative methods share the request policy and intervention budget, whereas Takeover uses the same request condition but transfers the remaining generation to the LLM after the first request. Matched selector ablations additionally share logged states, top-8 SLM--LLM joint pools, architecture, and optimization budget. At test time, FutureBridge queries only next-token candidates, appends one selected token, and returns generation to the SLM. On held-out GSM8K, MATH-500, and OlympiadBench states, complete token-only candidate rollouts measure pairwise ranking accuracy and top-1 success only for diagnosis. The supplement provides filtering, LoRA, hardware, and cost details.

\subsection{Experimental Results}

Table~\ref{tab:main_results} compares FutureBridge with SLM-only and collaborative decoding methods using the same Qwen3 model pairs and benchmark protocol.

\paragraph{Overall comparison.}
With Qwen3-1.7B, FutureBridge obtains 50.38 Math Avg., improving over S2T and Takeover by 1.24 and 1.98 points. Compared with the strongest SLM-only method, S2T-Local, it gains 10.18 points, corresponding to a 25.3\% relative improvement. These results establish end-to-end effectiveness; the matched ablation and rollout analysis below test whether the gains are associated with receiver-conditioned future supervision.

\paragraph{Consistency across tasks and SLM scales.}
FutureBridge achieves the highest accuracy on every benchmark at both SLM scales. With Qwen3-0.6B, it obtains 43.64 Math Avg., improving over S2T by 7.16 points and over S2T-Local by 13.50 points. The consistent gains for the weaker receiver indicate that future-compatible token selection remains effective as SLM capacity decreases.

\begin{table}[!t]
\centering
\caption{Token-only inference efficiency for the primary Qwen3-1.7B SLM. LLM calls are averaged per problem, and latency is normalized to greedy SLM decoding. S2T and FutureBridge append one intervention token per call and generate no LLM suffix; Takeover is excluded because it delegates the remaining trajectory to the LLM.}
\label{tab:token_efficiency}
\footnotesize
\renewcommand{\arraystretch}{1.08}
\setlength{\tabcolsep}{3.6pt}
\begin{tabular*}{0.94\columnwidth}{@{\extracolsep{\fill}}lrrr}
\hline
Method & Math Avg. & Calls & Latency \\
\hline
Greedy SLM & 37.30 & 0.0 & $1.00\times$ \\
S2T & 49.14 & 4.8 & $3.18\times$ \\
FutureBridge & \textbf{50.38} & 4.8 & $3.30\times$ \\
\hline
\end{tabular*}
\end{table}

\paragraph{Accuracy under a matched collaboration budget.}
Table~\ref{tab:token_efficiency} isolates token-selection cost for the matched token-level methods. S2T and FutureBridge average 4.8 LLM calls and generate no suffix. FutureBridge gains 1.24 Math Avg. points over S2T with 3.8\% higher latency from joint-pool scoring. Takeover remains in Table~\ref{tab:main_results} as an end-to-end delegation baseline but is not treated as a token-only, budget-matched method.

\begin{figure*}[t]
\centering
\includegraphics[width=\textwidth]{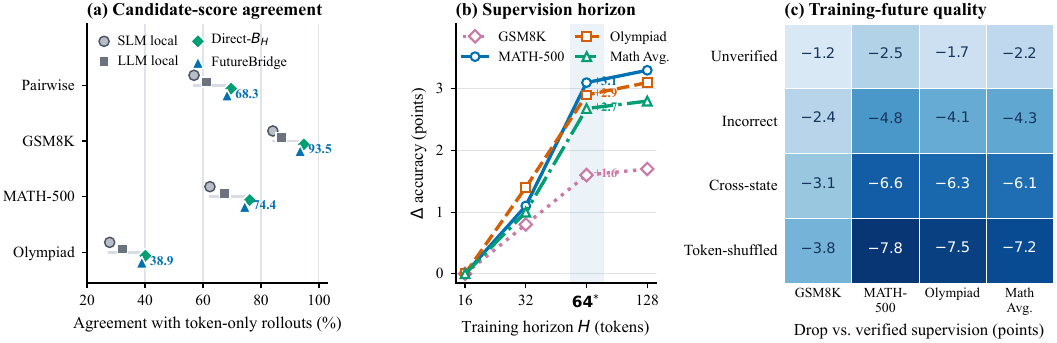}
\caption{Future-aware supervision improves agreement with token-only rollouts across GSM8K, MATH-500, and OlympiadBench, obtains most horizon gains by $H=64$, and degrades when the shared future is corrupted. (a) Candidate scores are evaluated against complete token-only SLM rollouts. (b) Extending the training horizon to $H=128$ provides only a further 0.1--0.2 accuracy points; changes are reported relative to $H=16$, and $64^{*}$ denotes the horizon selected on MATH validation. Deployment remains unchanged. (c) Each cell reports the accuracy drop relative to answer-verified future supervision.}
\label{fig:experimental_mechanisms}
\end{figure*}

\subsubsection{Candidate-Level Rollout Analysis}

Figure~\ref{fig:experimental_mechanisms}(a) compares each score with complete token-only candidate rollouts from matched states. Direct-$B_H$ uses the privileged shared future, whereas FutureBridge predicts its ranking from the state and candidate alone. Relative to LLM local scoring, Direct-$B_H$ and FutureBridge improve pairwise accuracy by 8.5 and 7.1 points. FutureBridge also raises top-1 rollout success by 6.5, 6.9, and 6.7 points on GSM8K, MATH-500, and OlympiadBench, remaining only 1.3, 1.7, and 1.2 points below Direct-$B_H$. Thus, the token-only selector preserves most of the future-aware signal across problem difficulty.

\subsection{Ablation Study}

\subsubsection{Candidate Pool and Supervision Target}

We first separate the effect of candidate coverage from the effect of the supervision target.

\begin{table}[!ht]
\centering
\caption{Candidate-pool and supervision-target ablations for Qwen3-1.7B (\%).}
\label{tab:component_ablation}
\scriptsize
\setlength{\tabcolsep}{2.3pt}
\begin{tabular*}{\columnwidth}{@{\extracolsep{\fill}}lrrrr}
\hline
Variant & GSM8K & MATH-500 & Olympiad & Math Avg. \\
\hline
FutureBridge & \textbf{96.8} & \textbf{81.6} & \textbf{43.5} & \textbf{50.38} \\
SLM-only pool & 94.4 & 78.7 & 41.2 & 47.9 \\
LLM-only pool & 94.9 & 79.4 & 41.8 & 48.3 \\
Direct LLM score & 93.5 & 78.6 & 40.5 & 47.2 \\
Distilled LLM pref. & 93.0 & 78.1 & 39.9 & 46.7 \\
\hline
\end{tabular*}
\end{table}

Table~\ref{tab:component_ablation} separates candidate availability from selection. SLM-only and LLM-only pools reduce Math Avg. by 2.5 and 2.1 points. Direct LLM scoring uses the same states, budget, and joint pool but ranks with $\log p_T(c\mid s_t)$; FutureBridge improves it by 3.3 points on GSM8K, 3.0 on MATH-500, 3.0 on OlympiadBench, and 3.2 in Math Avg. The distilled LLM-preference variant also matches the architecture and optimization budget, replacing only the target with Equation~\eqref{eq:hybrid_ablation} at $\alpha=0$; FutureBridge gains 3.8, 3.5, 3.6, and 3.7 points. Coverage and size-matched controls appear in the supplement.

\subsubsection{Shared-Future Horizon}

Figure~\ref{fig:experimental_mechanisms}(b) varies only the future used for offline targets. Extending $H$ from 64 to 128 still improves GSM8K, MATH-500, OlympiadBench, and Math Avg., but only by 0.1, 0.2, 0.2, and 0.1 points, respectively. Most of the gain is already obtained between $H=16$ and $H=64$, where the four metrics improve by 1.6, 3.1, 2.9, and 2.7 points. MATH validation selects $H=64$; deployment generates no future tokens.

\subsubsection{Shared-Future Quality}

Figure~\ref{fig:experimental_mechanisms}(c) changes only the offline training future. Unverified, incorrect, cross-state, and token-shuffled futures reduce GSM8K by 1.2, 2.4, 3.1, and 3.8 points and Math Avg. by 2.2, 4.3, 6.1, and 7.2 points. The same ordering holds on GSM8K, MATH-500, and OlympiadBench, indicating that effective supervision requires a state-specific, correctly ordered reasoning trajectory.

\FloatBarrier
\section{Conclusion}

In this work, we have formulated token-level small--large model collaboration as token reranking over a joint candidate pool. Rather than treating the LLM's local preference as the selection target, FutureBridge uses an answer-verified shared future to measure how well each candidate supports downstream reasoning under the receiving SLM. It fixes the reasoning state and future across candidates, derives candidate-level supervision from the frozen SLM, and distills this supervision into a deployable reranker that observes only the state and candidate token. The future is used only during training; at inference, FutureBridge selects and appends one token before returning generation to the SLM. This design separates candidate coverage from candidate selection and provides a controlled way to study the effects of candidate source, supervision horizon, and training-future quality. More broadly, token-level collaboration should select interventions according to whether the receiving SLM can use them to continue reasoning, rather than according to the LLM's local preference alone.

\clearpage
\bibliography{references}

\clearpage
\appendix
\setcounter{secnumdepth}{1}
\section*{Supplementary Material}

\section{Full Reproducibility Details}

\paragraph{Models and deployment boundary.}
The primary receiving model is Qwen3-1.7B, the scale-generalization receiver is
Qwen3-0.6B, and Qwen3-32B is the assisting LLM.  All base-model parameters
remain frozen.  Shared futures and complete candidate rollouts are used only
to construct training targets or evaluation diagnostics.  During deployment,
the LLM returns only its top-8 next-token candidates at an admitted state.
FutureBridge merges them with the SLM top-8, scores the deduplicated union,
appends exactly one selected token, and immediately returns decoding to the
SLM.  It neither requests nor appends an LLM reasoning suffix.

\paragraph{Request policy and collaboration budget.}
The external request policy computes the entropy of the SLM distribution on
its normalized top-64 support.  Its threshold is the 0.99 quantile estimated
on the problem-disjoint MATH validation split.  The threshold is frozen before
evaluation, and each trajectory permits at most eight admitted intervention
events.  The policy therefore uses no LLM output to decide whether to request
assistance.  Once a state is admitted, token-level collaborative methods
receive the same event and remaining intervention budget.  Takeover instead
delegates the remaining trajectory to the LLM at the first admitted state.
S2T ranks the SLM top-8 with LLM next-token probabilities, and FutureBridge
ranks the deduplicated SLM--LLM union.  Greedy decoding is used between
intervention events for the token-level methods, and all methods share the
same 4,096-token output limit, tokenizer, prompt, and answer verifier.

\paragraph{Data construction and filtering.}
We split the 7,500 MATH training problems into 6,750 training problems and 750
validation problems before collecting states, preventing prefixes from the
same problem from entering both subsets.  At every state admitted by the
frozen request policy, Qwen3-32B greedily produces one complete trajectory.
We discard a group when the final answer is incorrect or unparsable, the
post-action trajectory contains fewer than 128 tokens, or tokenization and
deduplication leave an invalid candidate group.  Table~\ref{tab:data_filter}
reports the resulting data volume.  The overall retained fractions are
58.2\% for Qwen3-1.7B states and 58.3\% for Qwen3-0.6B states.

\begin{center}
\begin{minipage}{\columnwidth}
\centering
\captionof{table}{Training-data construction and filtering by receiving SLM.}
\label{tab:data_filter}
\small
\setlength{\tabcolsep}{4.0pt}
\begin{tabular}{lrr}
\hline
Stage & 1.7B & 0.6B \\
\hline
Logged admitted states & 31,680 & 34,560 \\
Answer-verified trajectory & 21,216 & 23,424 \\
At least 128 future tokens & 18,912 & 20,736 \\
Valid joint candidate group & 18,432 & 20,160 \\
\quad Training groups & 16,576 & 18,144 \\
\quad Validation groups & 1,856 & 2,016 \\
\hline
\end{tabular}
\end{minipage}
\end{center}

\paragraph{LoRA reranker.}
The candidate scorer is initialized from the corresponding receiving SLM.
LoRA adapters are applied to the query, key, value, and output projections of
each self-attention block with rank 16, scaling 32, and dropout 0.05.  A linear
scalar head reads the hidden state of the appended candidate token.  Only the
LoRA parameters and scalar head are optimized; embeddings, transformer
weights, and the language-model head remain frozen.  We use bfloat16,
AdamW with learning rate $2\times10^{-4}$, weight decay 0.01,
$\beta_1=0.9$, $\beta_2=0.999$, a 3\% linear warmup, and gradient clipping at
1.0.  One candidate group forms a microbatch, gradients are accumulated over
32 groups, and training lasts three epochs.  The target temperature is
$\tau=0.5$.  We train one scorer for each
$H\in\{16,32,64,128\}$ under identical budgets and select $H=64$ on MATH
validation.  All reported FutureBridge results use three training seeds; the
main-table Math Avg.\ reports the seed mean without per-cell standard
deviations.

\paragraph{Hardware and software.}
Data construction, compatibility scoring, reranker training, and evaluation
use NVIDIA H200 GPUs with 141\,GB memory.  Each timing run uses one GPU and
batch size one so that latency is comparable across methods; independent seeds
may run concurrently but are timed separately.  Models run without
quantization under CUDA 12.4, PyTorch 2.6, and Transformers 4.52.  We reuse
the SLM key--value cache up to the admitted state and batch the candidate
branches for both offline compatibility scoring and deployed reranking.

\section{Cost Accounting}

\paragraph{Offline supervision cost.}
Table~\ref{tab:offline_cost} separates the one-time construction and training
cost for the primary Qwen3-1.7B receiver.  Generating answer-verified shared
futures is the largest component.  Compatibility scores for all four horizons
are obtained from the same length-128 likelihood matrix, so shorter horizons
do not require additional SLM forward passes.  The twelve reranker runs
correspond to four horizons and three random seeds.

\begin{center}
\begin{minipage}{\columnwidth}
\centering
\captionof{table}{One-time offline cost for the primary receiver.}
\label{tab:offline_cost}
\small
\setlength{\tabcolsep}{4.0pt}
\begin{tabular}{lrr}
\hline
Stage & Units & H200 hours \\
\hline
SLM state logging & 31,680 states & 5.8 \\
Verified-future generation & 21,216 trajectories & 29.4 \\
SLM compatibility scoring & 18,432 groups & 8.6 \\
Reranker optimization & 12 runs & 7.2 \\
\hline
Total & --- & 51.0 \\
\hline
\end{tabular}
\end{minipage}
\end{center}

\paragraph{Deployment cost.}
Table~\ref{tab:deployment_cost} reports token-only inference for S2T and
FutureBridge under the matched request policy.  ``Pool'' is the mean number of tokens considered at an
intervention event, ``appended'' counts the single intervention token, and
latency is normalized to greedy SLM decoding.  An LLM call returns next-token
logits only.  Thus, both token-level methods append one intervention token
per admitted event and generate zero LLM suffix tokens.  FutureBridge's
additional latency and memory arise from batching 13.2 candidate-scoring
branches through the LoRA reranker, not from generating future reasoning.
Takeover is excluded because its LLM generates the remaining trajectory.

\begin{center}
\begin{minipage}{\columnwidth}
\centering
\captionof{table}{Per-problem deployment cost for Qwen3-1.7B.}
\label{tab:deployment_cost}
\small
\setlength{\tabcolsep}{2.8pt}
\begin{tabular}{lrrrrrr}
\hline
Method & Calls & Pool & Appended & Suffix & Latency & GB \\
\hline
Greedy SLM & 0.0 & --- & 0.0 & 0 & $1.00\times$ & 8.6 \\
S2T & 4.8 & 8.0 & 4.8 & 0 & $3.18\times$ & 72.7 \\
FutureBridge & 4.8 & 13.2 & 4.8 & 0 & $3.30\times$ & 78.1 \\
\hline
\end{tabular}
\end{minipage}
\end{center}

\section{Additional Baseline Results}

Tables~\ref{tab:additional_baselines_17b} and
\ref{tab:additional_baselines_06b} report the baselines omitted from the
focused main comparison under the same Qwen3 protocol.

\begin{center}
\begin{minipage}{\columnwidth}
\centering
\captionof{table}{Additional baseline accuracy with Qwen3-1.7B (\%).}
\label{tab:additional_baselines_17b}
\scriptsize
\setlength{\tabcolsep}{2.5pt}
\begin{tabular*}{\columnwidth}{@{\extracolsep{\fill}}lrrrr}
\hline
Benchmark & TSD-KD & Mul-T & TaH & SpecR \\
\hline
GSM8K         & 75.3 & 77.1 & 79.0 & 89.2 \\
MATH-500      & 56.1 & 57.8 & 59.6 & 72.9 \\
OlympiadBench & 21.0 & 22.7 & 23.6 & 40.6 \\
AIME 2024     & 3.3  & 4.4  & 4.4  & 11.1 \\
AIME 2025     & 2.2  & 2.2  & 3.3  & 10.0 \\
Math Avg.     & 31.6 & 32.8 & 34.0 & 44.8 \\
\hline
\end{tabular*}
\end{minipage}
\end{center}

\begin{center}
\begin{minipage}{\columnwidth}
\centering
\captionof{table}{Additional baseline accuracy with Qwen3-0.6B (\%).}
\label{tab:additional_baselines_06b}
\scriptsize
\setlength{\tabcolsep}{2.5pt}
\begin{tabular*}{\columnwidth}{@{\extracolsep{\fill}}lrrrr}
\hline
Benchmark & TSD-KD & Mul-T & TaH & SpecR \\
\hline
GSM8K         & 55.7 & 57.6 & 58.9 & 67.9 \\
MATH-500      & 31.4 & 32.8 & 34.6 & 44.8 \\
OlympiadBench & 10.5 & 11.8 & 12.1 & 21.6 \\
AIME 2024     & 1.1  & 2.2  & 2.2  & 5.6 \\
AIME 2025     & 1.1  & 1.1  & 1.1  & 4.4 \\
Math Avg.     & 20.0 & 21.1 & 21.8 & 28.9 \\
\hline
\end{tabular*}
\end{minipage}
\end{center}

\section{Detailed Mechanism Results}

\paragraph{Candidate-level rollout agreement.}
Table~\ref{tab:rollout_detail} gives the numerical results underlying the
candidate-level analysis in the main paper.  Each held-out candidate is
appended alone, after which the frozen SLM completes the trajectory without
an LLM suffix.  Direct-$B_H$ is a privileged training-time score rather than
an inference oracle.  FutureBridge preserves most of its advantage over the
LLM local score while observing only the current state and candidate.

\begin{center}
\begin{minipage}{\columnwidth}
\centering
\captionof{table}{Agreement with complete token-only candidate rollouts (\%).}
\label{tab:rollout_detail}
\scriptsize
\setlength{\tabcolsep}{2.2pt}
\begin{tabular}{lrrrr}
\hline
Score & Pairwise & GSM8K & MATH-500 & Olympiad \\
& accuracy & top-1 success & top-1 success & top-1 success \\
\hline
SLM local score & 56.9 & 84.1 & 62.4 & 27.8 \\
LLM local score & 61.2 & 87.0 & 67.5 & 32.2 \\
Direct-$B_H$ & \textbf{69.7} & \textbf{94.8} & \textbf{76.1} & \textbf{40.1} \\
FutureBridge & 68.3 & 93.5 & 74.4 & 38.9 \\
\hline
\end{tabular}
\end{minipage}
\end{center}

\paragraph{Candidate-pool coverage.}
Table~\ref{tab:candidate_coverage} measures the fraction of matched held-out
states whose pool contains a token yielding a correct complete SLM rollout.
The top-8 sets overlap by 2.8 tokens on average, producing 13.2 joint
candidates and improving coverage over the stronger LLM-only pool by 9.3
points on MATH-500 and 8.8 points on OlympiadBench. On GSM8K, the
corresponding gain is 8.8 points.

\begin{center}
\begin{minipage}{\columnwidth}
\centering
\captionof{table}{Candidate coverage at matched held-out intervention states (\%).}
\label{tab:candidate_coverage}
\scriptsize
\setlength{\tabcolsep}{2.8pt}
\begin{tabular}{lrrrr}
\hline
Pool & Avg. size & GSM8K & MATH-500 & Olympiad \\
\hline
SLM top-8 & 8.0 & 72.8 & 49.6 & 45.2 \\
LLM top-8 & 8.0 & 79.4 & 54.1 & 50.7 \\
Joint pool & \textbf{13.2} & \textbf{88.2} & \textbf{63.4} & \textbf{59.5} \\
\hline
\end{tabular}
\end{minipage}
\end{center}

A size-matched SLM top-16 pool obtains 94.6 on GSM8K, 79.0 on MATH-500,
41.5 on OlympiadBench, and 48.0 Math Avg., remaining 2.4 Math Avg. points
below the joint pool.

\paragraph{Future horizon and quality.}
Table~\ref{tab:horizon_detail} reports the horizon sweep, where the future
length changes only the offline target.  Table~\ref{tab:quality_detail}
changes the shared future used for training while preserving the same logged
states, joint candidate pools, model architecture, and deployment procedure.
Both studies select and append only one token at inference.

\begin{center}
\begin{minipage}{\columnwidth}
\centering
\captionof{table}{Shared-future horizon used for offline supervision (\%).}
\label{tab:horizon_detail}
\small
\setlength{\tabcolsep}{3.0pt}
\begin{tabular}{lrrrr}
\hline
$H$ & GSM8K & MATH-500 & Olympiad & Math Avg. \\
\hline
16 & 95.2 & 78.5 & 40.6 & 47.7 \\
32 & 96.0 & 79.6 & 42.0 & 48.7 \\
64 & 96.8 & 81.6 & 43.5 & 50.38 \\
128 & \textbf{96.9} & \textbf{81.8} & \textbf{43.7} & \textbf{50.5} \\
\hline
\end{tabular}
\end{minipage}
\end{center}

\begin{center}
\begin{minipage}{\columnwidth}
\centering
\captionof{table}{Quality of the future used to construct training targets (\%).}
\label{tab:quality_detail}
\scriptsize
\setlength{\tabcolsep}{2.5pt}
\begin{tabular}{lrrrr}
\hline
Training future & GSM8K & MATH-500 & Olympiad & Math Avg. \\
\hline
Answer-verified & \textbf{96.8} & \textbf{81.6} & \textbf{43.5} & \textbf{50.38} \\
Unverified & 95.6 & 79.1 & 41.8 & 48.2 \\
Incorrect answer & 94.4 & 76.8 & 39.4 & 46.1 \\
Cross-state & 93.7 & 75.0 & 37.2 & 44.3 \\
Token-shuffled & 93.0 & 73.8 & 36.0 & 43.2 \\
\hline
\end{tabular}
\end{minipage}
\end{center}

\end{document}